# scientific reports

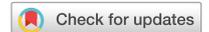

# OPEN  Random forest-based prediction of stroke outcome

Carlos Fernandez-Lozano[1,2], Pablo Hervella[3], Virginia Mato-Abad[4], Manuel Rodríguez-Yáñez[5], Sonia Suárez-Garaboa[4], Iria López-Dequidt[5], Ana Estany-Gestal[6], Tomás Sobrino[3], Francisco Campos[3], José Castillo[3], Santiago Rodríguez-Yáñez[4]✉ & Ramón Iglesias-Rey[3]✉

We research into the clinical, biochemical and neuroimaging factors associated with the outcome of stroke patients to generate a predictive model using machine learning techniques for prediction of mortality and morbidity 3-months after admission. The dataset consisted of patients with ischemic stroke (IS) and non-traumatic intracerebral hemorrhage (ICH) admitted to Stroke Unit of a European Tertiary Hospital prospectively registered. We identified the main variables for machine learning Random Forest (RF), generating a predictive model that can estimate patient mortality/ morbidity according to the following groups: (1) IS + ICH, (2) IS, and (3) ICH. A total of 6022 patients were included: 4922 (mean age 71.9 ± 13.8 years) with IS and 1100 (mean age 73.3 ± 13.1 years) with ICH. NIHSS at 24, 48 h and axillary temperature at admission were the most important variables to consider for evolution of patients at 3-months. IS + ICH group was the most stable for mortality prediction [0.904 ± 0.025 of area under the receiver operating characteristics curve (AUC)]. IS group presented similar results, although variability between experiments was slightly higher (0.909 ± 0.032 of AUC). ICH group was the one in which RF had more problems to make adequate predictions (0.9837 vs. 0.7104 of AUC). There were no major differences between IS and IS + ICH groups according to morbidity prediction (0.738 and 0.755 of AUC) but, after checking normality with a Shapiro Wilk test with the null hypothesis that the data follow a normal distribution, it was rejected with W = 0.93546 (p-value < 2.2e−16). Conditions required for a parametric test do not hold, and we performed a paired Wilcoxon Test assuming the null hypothesis that all the groups have the same performance. The null hypothesis was rejected with a value < 2.2e−16, so there are statistical differences between IS and ICH groups. In conclusion, machine learning algorithms RF can be effectively used in stroke patients for long-term outcome prediction of mortality and morbidity.

Stroke is the second leading cause of death and the third leading cause of disability in the world. Approximately 15 million people will experience a stroke episode every year worldwide of which 33% is left with a permanent disability, whereas 40% of the cases will result in death, and by 2030 will result in the annual loss of over 200 million (death or disability) globally[1]. Developing an appropriate long-term management plan and studying the progress in the management of stroke patients is necessary in order to organize healthcare structures for the coming years[1–5]. Predicting functional outcome after stroke would help clinicians make patient-specific decisions[6–10].

Machine learning (ML) provides a promising tool for disease evolution prediction and it is being increasingly used in biomedical studies. The application of ML in healthcare is widely anticipated as a key step towards improving care quality, and would play a fundamental role in the development of learning healthcare systems. Large scale studies in the general literature provide evidence in favor of some classier families such as Random

[1]Department of Computer Science and Information Technologies, Faculty of Computer Science, CITIC-Research Center of Information and Communication Technologies, Universidade da Coruña, A Coruña, Spain. [2]Grupo de Redes de Neuronas Artificiales y Sistemas Adaptativos. Imagen Médica y Diagnóstico Radiológico (RNASA-IMEDIR). Instituto de Investigación Biomédica de A Coruña (INIBIC). Complexo Hospitalario Universitario de A Coruña (CHUAC), SERGAS, Universidade da Coruña, A Coruña, Spain. [3]Clinical Neurosciences Research Laboratory (LINC), Health Research Institute of Santiago de Compostela (IDIS), Santiago de Compostela, Spain. [4]Software Engineering Laboratory, Department of Computer Science and Information Technologies, Faculty of Computer Science, University of A Coruña, Campus de Elviña, 15071 A Coruña, Spain. [5]Stroke Unit, Department of Neurology, Health Research Institute of Santiago de Compostela (IDIS), Hospital Clínico Universitario, Rúa Travesa da Choupana, s/n, 15706 Santiago de Compostela, Spain. [6]Unit of Methodology of the Research, Health Research Institute of Santiago de Compostela (IDIS), Santiago de Compostela, Spain. ✉email: santiago.rodriguez@udc.es; ramon.iglesias.rey@sergas.es

| | IS + ICH (n = 6022) | IS (n = 4922) | ICH (n = 1100) |
|---|---|---|---|
| **Demographic variables** | | | |
| Age (years) | 72.1 ± 13.7 | 71.9 ± 13.8 | 73.3 ± 13.1 |
| Female gender (%) | 44.5 | 44.8 | 41.5 |
| Arterial hypertension (%) | 66.7 | 63.7 | 60.7 |
| Diabetes mellitus (%) | 23.4 | 24.1 | 20.4 |
| Alcohol consumption (%) | 12.2 | 11.5 | 15.4 |
| Smoking (%) | 15.4 | 16.4 | 10.7 |
| Dyslipidemia (%) | 35.4 | 35.1 | 36.7 |
| Peripheral arterial disease (%) | 5.7 | 5.9 | 4.6 |
| Ischemic heart disease (%) | 10.8 | 11.3 | 8.6 |
| Atrial fibrillation (%) | 20.8 | 24.1 | 18.1 |
| Previous transient ischemic attack (%) | 5.4 | 6.1 | 2.5 |
| Previous ischemic stroke (%) | 13.1 | 13.6 | 9.8 |
| Previous intracerebral hemorrhage,% | 2.4 | 0.9 | 9.8 |
| Previous anticoagulants (%) | 9.5 | 8.5 | 14.1 |
| Previous platelet antiaggregants (%) | 22.9 | 24.4 | 16.5 |

**Table 1.** Demographic variables of the experimented dataset of patients summarized by group.

Forest (RF) in terms of classification performance[11,12]. RF has recently been used successfully in a wide range of biomedical applications, such as the automatic detection of pulse during electrocardiogram-based cardiopulmonary resuscitation or in breast cancer diagnosis using mammography images[13–18].

As to stroke, most studies focus on the use of ML methods to detect ischemic stroke (IS) lesions using neuroimaging data[19–23] and outcome estimation[24–28]. It has only been recently, however, that a study evaluated stroke outcome prediction at 3 months also in a group of non-traumatic intracerebral hemorrhage (ICH) patients using a nationwide disease registry[27]. Previous studies concluded that ML techniques can be effective to predict functional outcome of IS long-term patients or for prediction of symptomatic intracranial haermorrahe following thrombolysis from CT images. However, all works agree on the need to carry out further studies in order to confirm results, incorporate new variables and resolve their limitations or weaknesses.

Taking into account the prevalence of cerebrovascular diseases, accurately predicting stroke evolution is essential to stratify the rehabilitation care that should be administered, especially to patients with the best chance of recovery. The administration of rehabilitative therapies to those who are unlikely to benefit from them is inefficient for the Healthcare System and inconvenient and unproductive for patients. A predictive model that identifies stroke patients at risk of deterioration would make it possible to select/follow-up patients for reperfusion treatments, and increase the control of therapeutic homeostasis, thus addressing the needs of each patient individually. Furthermore, regarding to new regenerative cellular or molecular therapies, it is essential to identify the most suitable patients to respond accurately to treatments. We hypothesized that models developed with ML techniques based on the demographic, clinical, biochemical and neuroimaging variables obtained in the first 48 h after stroke are accurate stroke mortality and morbidity predictors at 3 months.

## Results

We included in the study 6022 patients; 4922 (81.8%) presented with IS and 1100 (18.2%) with ICH. We excluded 228 patients, who died during the first 24 h, and 84 with no follow-up at 3 months. The 65 features of the different groups included in the experimented dataset are shown in Tables 1, 2 and 3. Figure 1 lists flowchart of patient groups with their functional outcome and divided into morbidity and mortality.

Of the 4922 IS patients valid for this study, 55.2% were male and 44.8% female; the mean age was 71.9 ± 13.8 years. According to the TOAST classification, 1127 patients were classified as atherothrombotic (22.9%), 1786 as cardioembolic (36.3%), 428 as lacunar (8.7%) and 1520 as undetermined (30.9%). Poor functional outcome at 3 months was found in 47.5% of IS patients, thus showing a morbidity of 33.4% and a mortality of 13.2%.

Of the 1100 ICH patients, 58.5% were male and 41.5% female; the mean age was 73.3 ± 13.1 years. ICH etiology was related with 506 hypertensive patients (46%), 114 with amyloid angiopathy (10.4%), with 156 anticoagulants (14.2%) and 323 with other causes (29.4%). 58.6% of ICH patients showed poor outcome at 3 months, with a morbidity of 27.6% and a mortality of 30.2%.

Using the filter feature selection, the dataset was reduced to only 7 variables: National Institute of Health Stroke Scale score at admission [NIHSS (0)]; NIHSS score at 24 h [NIHSS (24)]; NIHSS score at 48 h [NIHSS (48)]; Axillary temperature at admission [T(0)]; Early neurological deterioration [ED]; Leukocytes at admission [LEU (0)]; and blood glucose at admission [GLU (0)] as with these variables, RF was much more stable and deviations or variations between experiments could be reduced.

|  | IS + ICH (n = 6022) | IS (n = 4922) | ICH (n = 1100) |
|---|---|---|---|
| **Clinical/Neuroimaging variables** | | | |
| Stroke on awakening (%) | 8.3 | 9.1 | 4.6 |
| Previous mRS | 0 [0, 1] | 0 [0, 1] | 1 [0, 1] |
| Time from stroke onset, minutes | 239.1 ± 175.2 | 240.8 ± 167.4 | 231.3 ± 206.1 |
| NIHSS score at admission | 13 [7, 19] | 13 [8, 19] | 13 [7, 18] |
| NIHSS score at 24 h | 8 [13, 16] | 7 [3, 15] | 12 [6, 19] |
| NIHSS score at 48 h | 7 [2, 15] | 6 [2, 14] | 12 [4, 20] |
| Early neurological deterioration (%) | 7.7 | 5.8 | 16.5 |
| **TOAST** | | | |
| Atherothrombotic (%) | – | 22.9 | – |
| Cardioembolic (%) | – | 36.3 | – |
| Lacunar (%) | – | 8.7 | – |
| Undetermined (%) | – | 30.9 | – |
| Others (%) | – | 1.2 | – |
| Intravenous fibrinolysis (%) | – | 22.7 | – |
| Th ombectomy (%) | – | 5.2 | – |
| DWI at admission (ml) | – | 33.3 ± 76.9 | – |
| TC volume 4th–7th day (ml) | – | 51.1 ± 82.3 | – |
| **Hemorrhagic transformation of IS** | | | |
| IH1 (%) | – | 7.0 | – |
| IH2 (%) | – | 3.1 | – |
| PH1 (%) | – | 1.7 | – |
| PH2 (%) | – | 1.2 | – |
| **Etiology of ICH** | | | |
| Hypertensive (%) | – | – | 46.0 |
| Amyloid (%) | – | – | 10.4 |
| Anticoagulants (%) | – | – | 14.2 |
| Others/Undetermined (%) | – | – | 29.4 |
| Hematoma volume at admission (ml) | – | – | 40.3 ± 46.2 |
| Hematoma volume 4th–7th day (ml) | – | – | 51.9 ± 48.1 |
| Total hematoma volume (ml) | – | – | 68.3 ± 53.1 |
| Volume of hypodensity (ml) | – | – | 15.2 ± 17.9 |
| Hematoma growth (ml) | – | – | 11.9 ± 27.6 |
| **Topography** | | | |
| Deep hemispherics (%) | – | – | 50.0 |
| Lobar (%) | – | – | 39.6 |
| Cerebellar (%) | – | – | 4.7 |
| Breinstem (%) | – | – | 3.8 |
| Primary intraventricular (%) | – | – | 1.9 |
| Axillary temperature at admission (ºC) | 36.4 ± 0.7 | 36.4 ± 0.7 | 36.6 ± 0.8 |
| Blood glucose at admission (mg/dl) | 137.6 ± 56.3 | 137.3 ± 57.9 | 138.9 ± 48.1 |
| Sedimentation rate (mm) | 26.4 ± 23.1 | 26.5 ± 23.1 | 26.2 ± 23.1 |
| Glycosylated hemoglobin (%) | 6.1 ± 2.1 | 6.1 ± 2.3 | 5.8 ± 0.9 |
| LDL cholesterol (mg/dl) | 101.9 ± 42.9 | 112.5 ± 44.4 | 109.6 ± 35.2 |
| HDL cholesterol (mg/dl) | 41.2 ± 18.5 | 41.8 ± 18.5 | 38.8 ± 18.3 |
| Triglycerides (mg/dl) | 118.3 ± 63.1 | 121.2 ± 65.1 | 109.4 ± 50.7 |
| Platelets (× $10^3$/ml) | 215.4 ± 82.9 | 217.7 ± 83.7 | 203.3 ± 77.9 |
| Hemoglobin (g/dl) | 13.7 ± 1.9 | 13.8 ± 1.9 | 13.5 ± 2.1 |
| DBP at admission (mmHg) | 81.9 ± 16.1 | 81.5 ± 15.8 | 84.3 ± 17.2 |
| SBP at admission (mmHg) | 152.9 ± 27.3 | 152.5 ± 27.3 | 155.5 ± 27.4 |

**Table 2.** Clinical and neuroimaging variables of the experimented dataset of patients summarized by group.

**Prediction of mortality.** Figure 2A shows the most important variables for the model associated to IS, ICH or IS + ICH patient groups in relation to the mortality prediction. The value shown is, in all cases, the sum of the importance obtained by the algorithm for the variable in each of the experiments internally.

|  | IS + ICH (n = 6022) | IS (n = 4922) | ICH (n = 1100) |
|---|---|---|---|
| **Molecular markers** | | | |
| Leukocytes at admission ($\times 10^3$/ml) | 8.9 ± 3.1 | 9.1 ± 3.2 | 8.8 ± 3.3 |
| Fibrinogen at admission (mg/dl) | 443.9 ± 101.7 | 444.5 ± 101.8 | 444.1 ± 101.5 |
| C-reactive protein admission (mg/dl) | 2.7 ± 3.8 | 3.6 ± 4.2 | 5.2 ± 5.2 |
| Microalbuminuria (mg/24 h) | 7.9 ± 26.2 | 5.9 ± 25.9 | 16.7 ± 30.0 |
| NT-pro-BNP levels (pg/ml) | 915.9 ± 1563.7 | 1581.2 ± 1886.1 | 1013.8 ± 3620.2 |
| **Outcome at 3 months** | | | |
| mRS | 2 [1, 4] | 2 [0, 4] | 3 [1, 6] |
| Poor outcome (%) | 49.6 | 47.5 | 58.6 |
| Morbidity (%) | 35.0 | 33.4 | 27.6 |
| Mortality (%) | 16.3 | 13.2 | 30.2 |

**Table 3.** Molecular markers and outcome at 3 months of the experimented dataset of patients summarized by group.

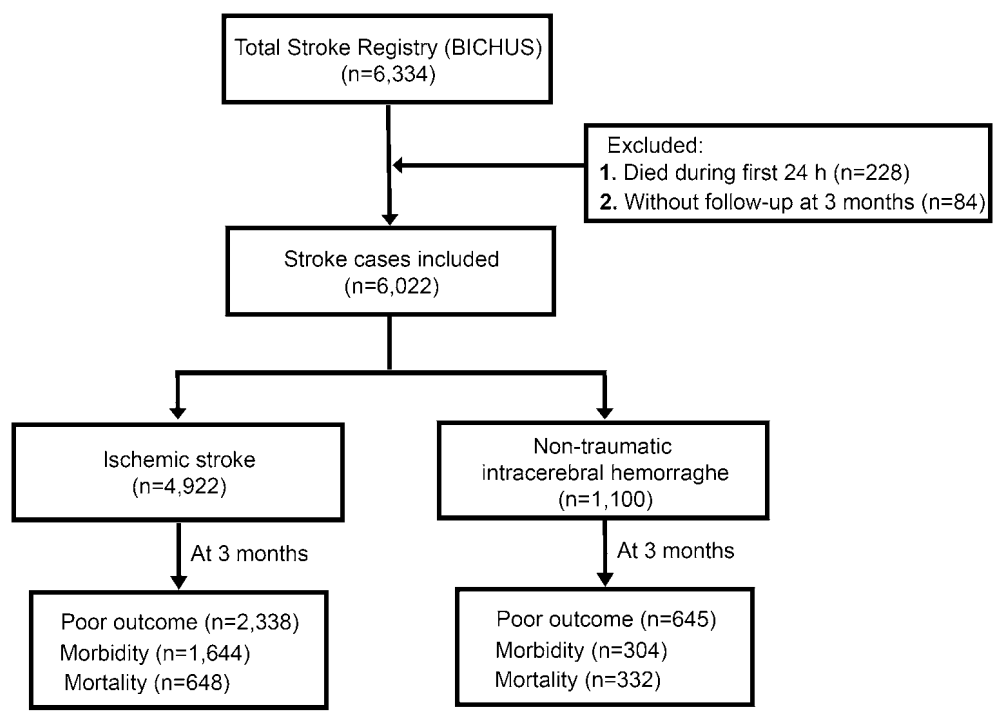

**Figure 1.** Flowchart of patient groups and functional outcome.

The most important variables, taking into account the three groups of patients analyzed, were NIHSS (48) and NIHSS (24). In the IS and ICH patient groups, the importance of ED, T (0) and NIHSS (0) should be highlighted. NIHSS (0) was also observed to be more important in patients with ICH than in those with IS when the models do not have data from both types of patients. It seems, however, that its importance is significantly reduced when the model has the complete set. Finally, LEU (0) and GLU (0) are variables that help balance the results for the complete model, reducing the variability of the individual IS or ICH models among all variables.

The variation obtained between all the repetitions performed in area under the receiver operating characteristics (ROC) curve (AUC) terms is detailed in Fig. 2B,C for the three experiments performed. The complete problem with two types of patients is the most stable with a minimum deviation between experiments (median of 0.904 ± 0.025 of AUC and 0.825 ± 0.030 of accuracy (ACC)). On the other hand, the ICH problem is the one in which RF has more problems to make adequate predictions as the range of results varies in more than 20 AUC points between the best (0.9837 of AUC with 0.94 of ACC) and the worst experiment (0.7104 of AUC and 0.6122 of ACC) and values for 100 repetitions of 0.875 ± 0.048 of AUC and 0.8 ± 0.052 of ACC. Th s prediction is therefore the most complex for the model.

As to the IS problem, RF presented similar values to those of the IS-ICH prediction problem although variability between experiments is slightly higher (0.909 ± 0.032 of AUC and 0.833 ± 0.040 of ACC). This led us to conclude that there is enough information within the selected variables so that when RF has enough patients in the dataset, the model predicts very accurately which patients are most likely to die on the basis of the data collected at admission. It was also observed that when there is a greater amount of data and patients are stratified into the three categories, the model is much more stable, and results are better. A 2D-heatmap of mortality predictions against NIHSS (48) and NIHSS (24) is detailed in Fig. 3 in order to explain the decision boundary of the model. Note that the misclassified items are highlighted and that the intensity of the colors also indicates the certainty of the prediction. We showed the IS + ICH group as it was the most stable for mortality prediction (0.904 ± 0.025 of AUC).

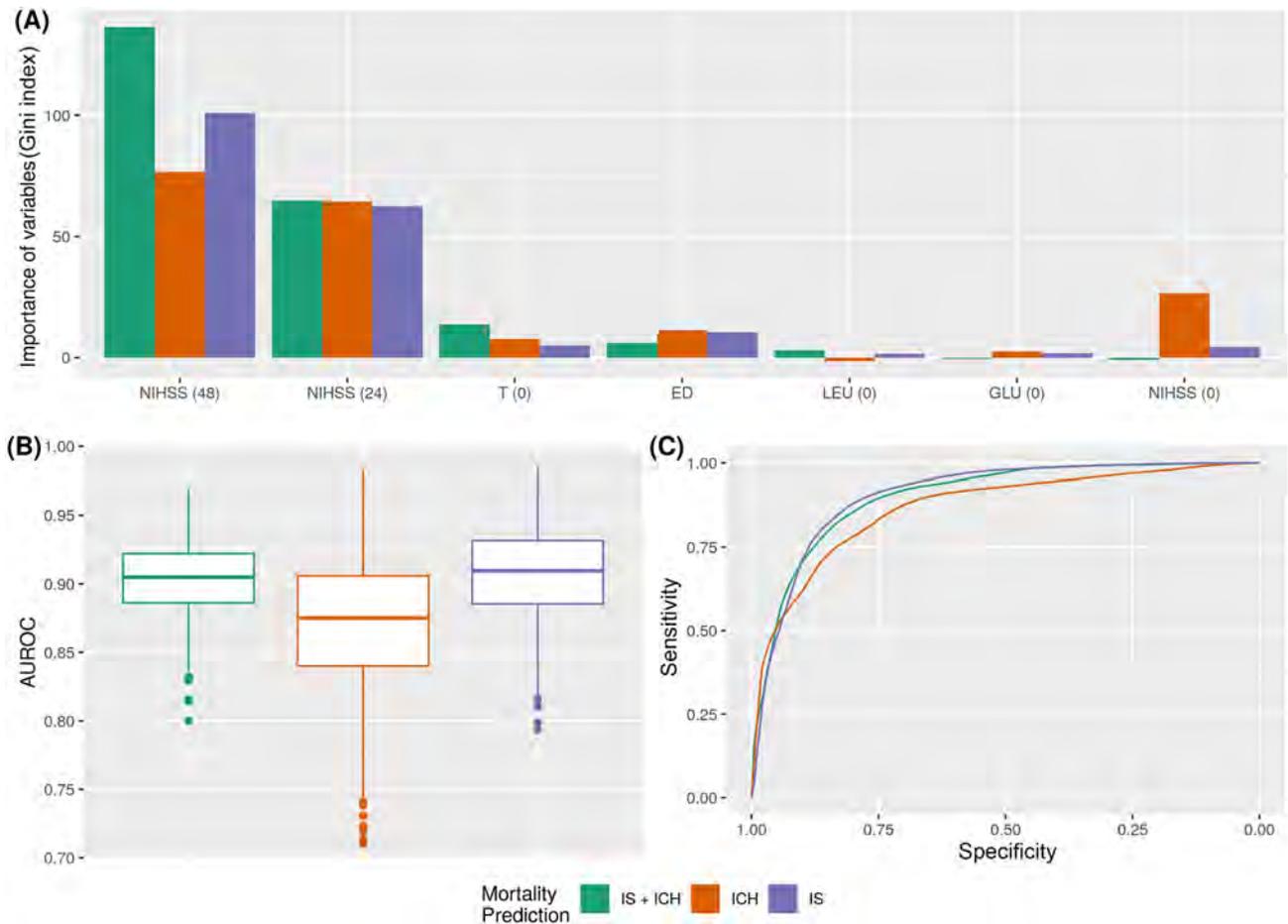

**Figure 2.** Mortality prediction for IS + ICH, IS and ICH groups. (**A**) Main variables for the machine learning model: NIHSS score at admission [NIHSS (0)]; NIHSS score at 24 h [NIHSS (24)]; NIHSS score at 48 h [NIHSS (48)]; Axillary temperature at admission [T(0)]; Early neurological deterioration [ED]; Leukocytes at admission [LEU (0)]; and Blood glucose at admission [GLU (0)]. (**B**) AUROC values obtained. (**C**) ROC curves for the Random Forest classifie .

**Prediction of morbidity.** Figure 4A shows that NIHSS (48) and NIHSS (24) are again the most important variables for the model associated to the three groups studied in relation to morbidity prediction. However, it seems that the variables ED for the IS and total groups, and GLU (0) for ICH patients provide relevant predictive capacity. NIHSS (0) was identifi d by the model as a variable with negative effects on classifi ation for the IS patient group as it worsens prediction. For the ICH patient group, however, NIHSS (0) is a key variable.

Figure 4B,C shows that there were no major differences between the IS and IS + ICH groups (0.738 and 0.755 of AUC and 0.683 and 0.700 of ACC) but with the ICH (0.667 of AUC and 0.618 of ACC).

**AUC with 7 input variables.** Figure 5 shows the comparison of ROC curves for the most important variables of the ML model associated to the groups IS + ICH, IS or ICH patients in relation to the mortality and morbidity prediction. IS + ICH and IS patient groups curves revealed NIHSS (24) and NIHSS (48) variables with the best AUC values obtained for both morbidity [0.677 (CI 95% 0.662–0.692) vs. 0.703 (CI 95% 0.686–0.719); and 0.669 (CI 95% 0.654–0.684) vs. 0.697 (CI 95% 0.680–0.713)] and mortality [0.888 (CI 95% 0.876–0.900) vs. 0.892 (CI 95% 0.878–0.906); and 0.897 (CI 95% 0.885–0.908) vs. 0.899 (CI 95% 0.885–0.913)] prediction.

The ICH group presented NIHSS (0) and NIHSS (24) with the best AUC values obtained for morbidity [0.591 (CI 95% 0.553–0.629) and 0.588 (CI 95% 0.550–0.626)]; and NIHSS (24) and NIHSS (48) for mortality [0.865 (CI 95% 0.838–0.891) and 0.873 (CI 95% 0.847–0.899)] estimation.

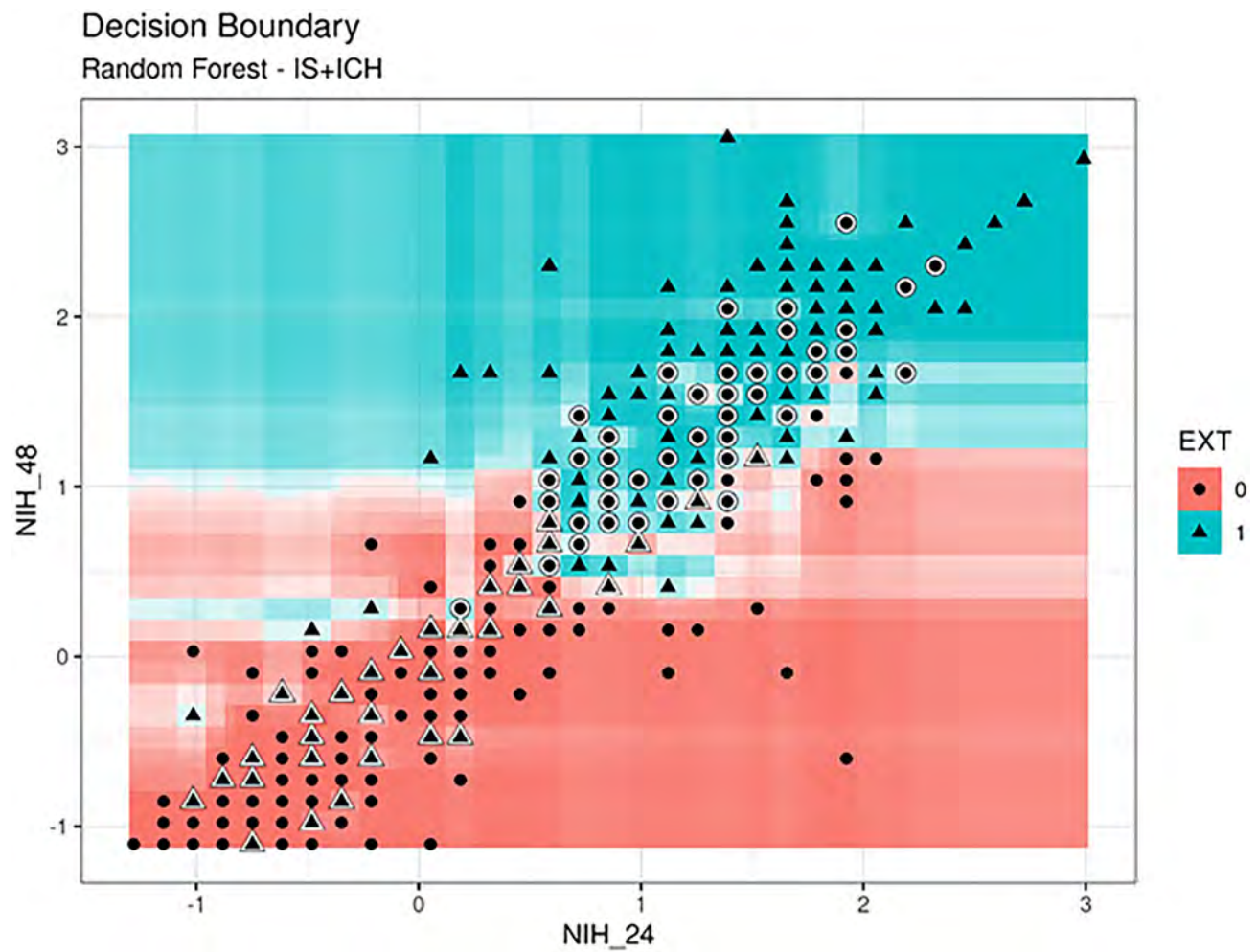

**Figure 3.** 2D-heatmap of mortality (EXT) predictions against NIHSS(48) and NIHSS(24). Model results are shown for the IS + ICH group, as it was the most stable for mortality prediction (0.904 ± 0.025 of AUC). Red areas correspond to patients who do not die (0), blue areas correspond to patients who die (1), and misclassified items are highlighted.

## Discussion

It is difficult but essential to accurately predict functional outcomes after stroke. Outcome prediction plays an important role in long-term decision making, patient treatment, organization of Health Centers, and domestic conditions. Plans could be developed on the basis of a better prediction of the degree of recovery of each patient with appropriate and individualized rehabilitation measures that take into account the domestic and economic conditions, leading to shared decisions with patients, relatives, and sociomedical centers[27,29].

The conventional approach to the evaluation of stroke outcomes data resorts to classical statistical models (logistic regression). Logistic regression models identify and validate predictive variables. Their main advantage is that they can be easily implemented and interpreted[24]. ML algorithms have the potential to outperform conventional regression because they are able to capture nonlinearities and complex interactions among multiple predictor variables. They can also handle large-scale multi-institutional data, with the added advantage of easily incorporating newly available data to improve prediction performance, and a better handling of a large number of predictors[30].

A recent systematic review found that ML was not superior to logistic regression in clinical prediction modeling[31]. However, inconsistent conclusions have been often found when comparing the performance of classical models to different machine learning algorithms in clinical applied studies. These studies agree that further research is needed to assess the feasibility and acceptance of ML applications in clinical practice[32,33]. Different research works have proposed strategies for stroke prediction based on ML algorithms with excellent results, but with a great diversity in the variables analyzed (clinical, molecular markers or imaging), calibration/training protocols performed, and models implemented (neural networks, tree-based and kernel-based methods). Some limitations of these previous studies are due to: (1) the low sample size used, (2) the characteristics of the patients evaluated; most of the studies only evaluate IS sub-groups, such as, IS patients treated with rTPA or endovascular intervention, (3) studies used demographic, clinical, molecular or neuroimaging variables independently and uncorrelated, and (4) the small number of variables used in ML models. Asadi et al. performed dichotomized modified Rankin Scale (mRS) models of acute ischemic stroke (n = 107) and presented 0.6 AUC of ANN and 70% accuracy of SVM[24]. Bentley et al. developed an SVM model to study acute ischemic stroke patients (n = 116) at risk for symptomatic intracranial hemorrhage using CT brain images[23]. Monterio et al. applied ML techniques (RF, Xgbosst, SVM and Decision tree) to predict the functional outcome of ischemic stroke patients (n = 425) treated with Recombinant Tissue Plasminogen Activator (rtPA) 3 months after the initial stroke. They started using only the information available at admission and then they went on to analyze how prediction improves by adding more features collected at different points in time after admission. The ML approach achieved AUC of 0.808 when using the features available at admission and as new features were progressively added, AUC increased to a value above 0.90[25]. Heo et al. researched into the applicability of machine learning-based models with a prospective cohort of 2604 patients with acute ischemic stroke. The AUC obtained was 0.888 for DNN model, 0.857 for RF model, and 0.849 for logistic regression model[26]. Recently, Lin et al. used a Taiwan Stroke Registry (n = 40,293) to evaluate several ML approaches (SVM, RF, ANN, and HANN) for 90-day stroke outcomes prediction. ML techniques presented over 0.94 AUC in both ischemic and hemorrhagic stroke using preadmission and inpatient data[27]. Alaka et al. concluded that both logistic regression and ML models had comparable predictive accuracy at 90 days when validated internally (AUC range = [0.65–0.72]) and externally (AUC range = [0.66–0.71]) in acute IS patients after endovascular treatment (n = 614–684)[28].

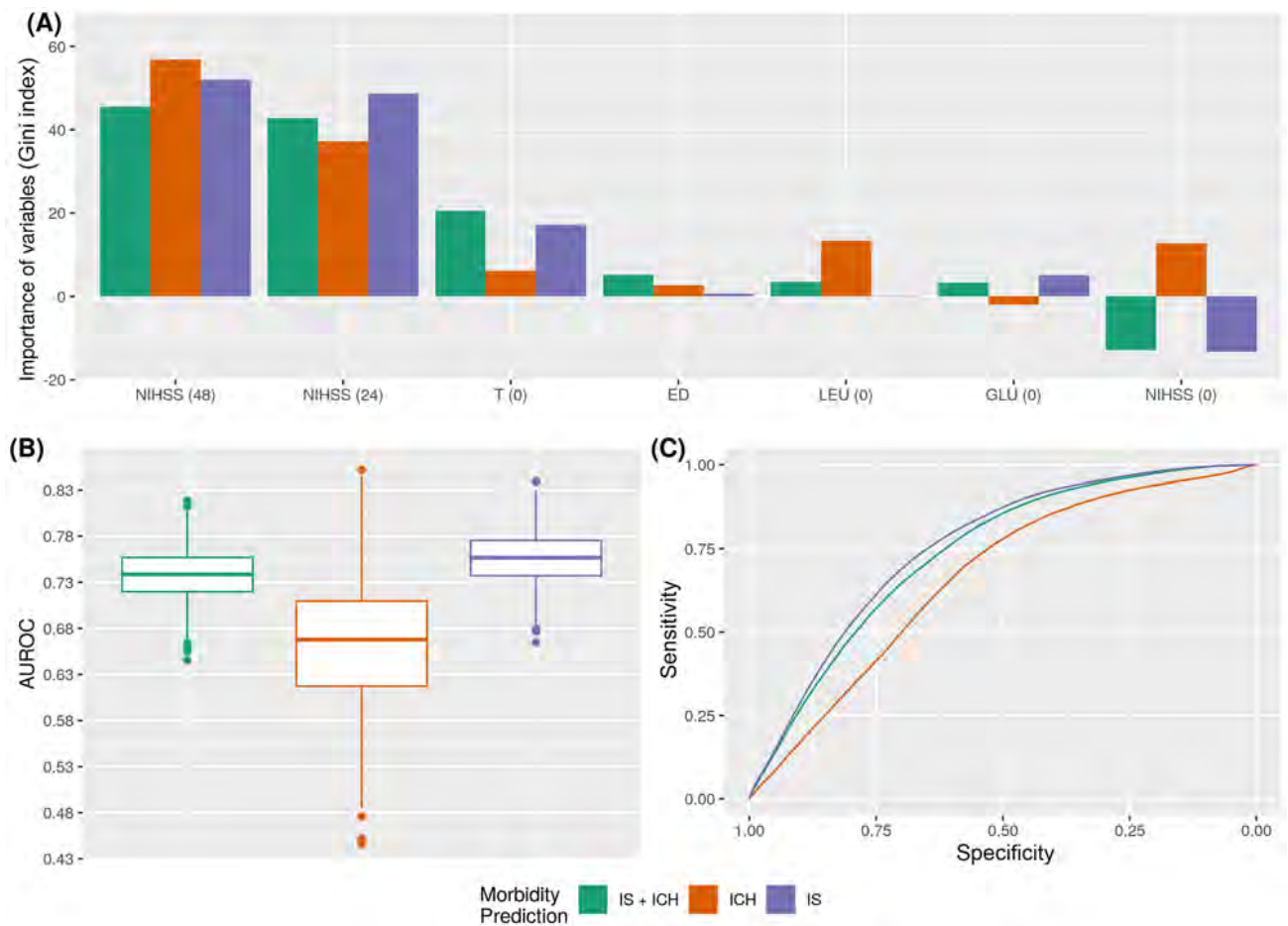

**Figure 4.** Morbidity prediction for IS + ICH, IS and ICH groups. (**A**) Main variables for the machine learning model: NIHSS score at admission [NIHSS (0)]; NIHSS score at 24 h [NIHSS (24)]; NIHSS score at 48 h [NIHSS (48)]; Axillary temperature at admission [T(0)]; Early neurological deterioration [ED]; Leukocytes at admission [LEU (0)]; and Blood glucose at admission [GLU (0)]. (**B**) AUROC values obtained. (**C**) ROC curves for the Random Forest classifie .

In our study, we analyzed a ML model of stroke prediction at 3 months using the Hospital's Stroke Registry (BICHUS) on the basis of demographic, clinical, molecular and neuroimaging variables. Mortality and morbidity were evaluated by identifying the main variables for the ML model. Data were studied as a whole (IS + ICH) or as independent subsets. Our ML classifie s exhibited high performance with over 0.90 AUC in the three groups evaluated in relation to the mortality outcome. The IS group had the best results (n = 4922). The model indicates that the most relevant variables are NIHSS (48) and NIHSS (24). In addition, the variable NIHSS (0) is also important for the ICH patients (n = 1100). The rest of the variables provide information marginally, although the importance of T(0) and ED should not be disregarded. On the other hand, AUC over 0.75 was found in the three groups evaluated in relation to the morbidity outcome. The model developed indicates that the most relevant variables are NIHSS (48) and NIHSS (24), although ED for the IS group and GLU (0) for the ICH provide predictive capability. Compared to ROC curve analysis of the 7 input variables, ML classifier has a high performance in three groups, with the NIHSS (0), NIHSS (24) or NIHSS (48) as the most influential predictors.

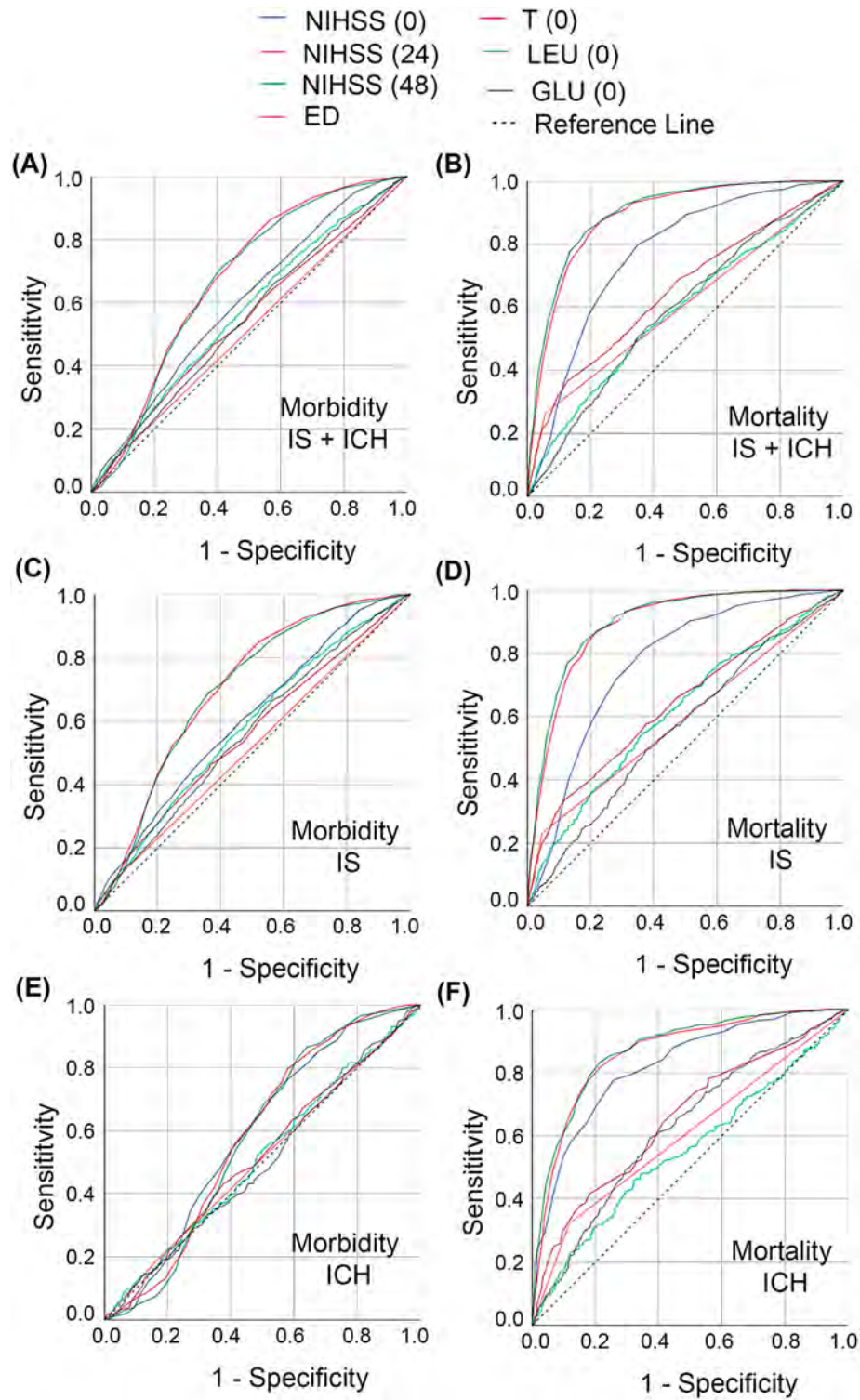

**Figure 5.** Comparison of ROC curves of 7 variables selected for machine learning experiments for mortality and morbidity prediction at 3 months of the different patient groups evaluated. (**A,B**) Morbidity and mortality of IS + ICH group. (**C,D**) Morbidity and mortality of IS group. (**E–F**) Morbidity and mortality of ICH group.

The findings in this report are subject to at least four limitations. First, this was a retrospective, single-center study with a relatively small clinical dataset. The intrinsic need for large training datasets may affect the accuracy of ML models in studies that could be overadjusted by irrelevant clinical predictors, or some predictors may be underestimated, thus increasing random errors. It is important to note that the variables selected in our study had been previously identified by means of a T-test and supervised by expert neurologists. However, in the future, both training and validation procedures will need to include a multicenter dataset and a prospective study to verify the model and the variables obtained. Second, the IS and the ICH patient groups were unbalanced. We consider, however, that the two types of stroke should be studied independently to find both differences and similarities. Third, we used RF machine learning algorithms, although other models like DNN or deep logistic regression could be used for comparison purposes. Four, we used typical clinical variables as inputs for the ML model, and we did not stratify patients in different subgroups, which could improve the results presented. We consider that it would be useful to evaluate the common variables from a clinical point of view, so once again emphasis is on the importance of NIHSS, axillary temperature and blood glucose. The major strengths of the present study include the large sample size (6022 patients; 4922 with IS and 1100 with ICH), which enabled study of the combination of different stroke types in detail (IS + ICH, IS, and ICH). Furthermore, to derive a global risk score for stroke, we have evaluated/interrelated demographic, clinical, biochemical and neuroimaging variables. Another distinctive feature of this analysis compared with previous studies is that we also included molecular markers associated with inflammation (leukocytes, fibrinogen and C-reactive protein), endothelial and atrial dysfunction (microalbuminuria and NT-proBNP).

## Conclusions

Machine learning algorithms, particularly Random Forest, can be effectively used in long-term outcome prediction of mortality and morbidity of stroke patients. NIHSS at 24, 48 h and axillary temperature are the most important variables to consider in the evolution of the patients at 3 months. Future studies could incorporate the use of imaging and genetic information. Furthermore, the robust model developed could be used in other applications and different scopes with similar data; such as traumatic brain injury, or dementia (Alzheimer's and Parkinson's disease).

## Materials and methods

**Patient selection.** The dataset used in this research work consisted of patients with IS and ICH admitted to the Stroke Unit of the Hospital Clínico Universitario of Santiago de Compostela (Spain), who were prospectively registered in an approved data bank (BICHUS). All patients were treated by a certified neurologist according to national and international guidelines. Exclusion criteria for this analysis were: (1) patients who died during the first 24 h, and (2) loss of follow-up (personal interview or telephone contact) at 3 months.

The analysis of the data for this study was retrospective, from September 2007 to September 2017. This research was carried out in accordance with the Declaration of Helsinki of the World Medical Association (2008) and approved by the Ethics Committee of Santiago de Compostela (2019/616). All patients or their relatives signed the informed consent for inclusion in the registry and for anonymous use of their personal data for research purposes.

**Demographic, clinical, molecular and neuroimaging variables.** The registry includes demographic variables, previous medical history and vital signs. Blood samples for hemogram, biochemistry and coagulation tests were obtained and analyzed at the central hospital laboratory. Neurological deficit was evaluated by a certified neurologist using the National Institute of Health Stroke Scale (NIHSS) at admission, and every 24/48 h during hospitalization. The modified Rankin Scale (mRS) was used to evaluate functional outcome at discharge and at 3 months[33,34].

Effective reperfusion of IS patients was defined as ≤ 8 points in the NIHSS during the first 24 h. Early neurological deterioration was defined as ≥ 4 points in NIHSS within the first 48 h with respect to baseline NIHSS score. Poor functional outcome was defined as mRS > 2 at 3 months, morbidity as 3 ≤ mRS ≤ 5, and mortality as mRS = 6. Ischemic stroke diagnosis was made using the TOAST criteria[35].

Computed Tomography (CT) was performed in all patients and Magnetic Resonance Imaging (MRI) in selected patients at admission. Follow-up CT scan after fibrinolysis or thrombectomy was performed in all IS patients at 24 h, and CT at 48 h or when neurological deterioration was detected and between the 4th–7th day. ICH and perihematomal edema volumes were calculated using the ABC/2 method[36]. ICH topography was classified as lobar when it predominantly affected the cortical/subcortical white matter of the cerebral lobes or as deep when it was limited to the internal capsule, the basal ganglia or the thalamus. All neuroimaging tests were analyzed by a neuro-radiologist supervised by the above certified neurologist.

**Outcome endpoints.** The objective of this research work was to identify the main predictors for the machine learning model in order to generate a predictive model using machine learning techniques for the prediction of mortality and morbidity of stroke patients according to their stratification to one of the following groups: 1) IS + ICH, 2) IS, and 3) ICH.

**Machine learning.** We used the RF algorithm for the prediction of mortality and morbidity of stroke patients. RF is an ensemble learning method, i.e., a strategy that aggregates many predictions to reduce the variance and improve the robustness and precision of outputs[37–39]. A remarkable characteristic of the RF is that it provides an internal measure of the relative importance of each feature on the prediction. This model generally works very well for any type of problem, regardless of size and even if the data are unbalanced or missing[37]. It also makes it possible to analyze the importance of the variables used by the model. To this end, the Gini importance index was calculated. This index measures the increase in impurity of each variable in the model when selected in the random distribution process. Each time a node selects a variable, the Gini impurity index for the two child nodes is lower than in the parent. It is not a simple final summation of the values obtained in all the trees for each variable but a weighting.

Generally speaking, all ML algorithms have a number of hyperparameters that must be optimized to obtain the best results for the particular problem they are analyzing. We used R[40,41] and the following packages: mlr[42] to calculate the best number of trees (ranging from 500 to 1000); Random Forest[43] or our experiments; and ggplot2[44] graphics for data analysis.

**Data pre-processing.** Balanced classes in classification problems are critical for ML algorithms. When analyzing the problem, we initially obtained prediction values in AUC lower than 0.65 in the best of cases using 65 features and four different ML algorithms, which is considered a bad performance value for prediction. This is mainly due to the fact that in our dataset there is a high percentage of patients who survived versus patients who died, and we found noise and correlation between features, confounding the predictors. These numbers show an unbalanced problem that needs to be addressed since a predictive model for patient death is being generated.

The following are the two main approaches to balance the data: (1) oversampling the minority class or (2) undersampling the class where the data has more examples. Although these are very powerful techniques that are able to increase the performance of the classifiers, they must be handled cautiously, more so in medical problems, to prevent overadjustments or the loss of generation capacity in the models when new synthetic samples are included (oversampling) at the learning phase of the algorithms[45]. In this work, undersampling methods (random undersampling from the majority class) were assessed for class balancing purposes. To ensure that the undersampling process is fair and that the generalization capability of the models is not biased we ran 100 repetitions, each with a different random undersampling, of a tenfold cross-validation experiment to observe the behavior and the stability of results. The more stability the better the random removing process. This means that the remaining samples of the majority class captured the underlying knowledge of the class.

The experimental design developed to analyze the original data included: a data preprocessing phase and the balancing of the subclasses; a tenfold cross-validation and 100 repetitions. For each of these repetitions, the position of each patient in the dataset was randomized. The preprocessed data were also randomized to avoid any potential process-related bias.

The problem was broken down into six different but complementary and informative problems: mortality and morbidity prediction with IS, ICH or IS + ICH patients. This approach sought to analyze more exhaustively the differences between the different types of patients when predicting death/poor outcome and to analyze whether the variables with more weight in the prediction were the same in all the cases.

In order to identify which of the 65 variables available are the most informative, we performed feature selection. There are mainly three different approaches for feature selection in machine learning: filter, wrapper and embedded[46]. Filter methods assess the relevance of each feature by looking only at the intrinsic properties of the data (independent of the algorithms). We calculated a feature relevance score (T-test) on the training data, and low-scoring features were removed choosing a manual cut-off point to reduce the variance of the models (see Supplementary Fig. S1 and Supplementary Material).

**Statistical analysis.** For the descriptive study of the quantitative variables, results were expressed as percentages for categorical variables and as mean (SD) or median (quartiles) for the continuous variables, depending on whether their distribution was normal or not. The Kolmogorov–Smirnov test was used for testing the normality of the distribution. To measure the performance of the model we used the area under the receiver operating characteristics (ROC) curve (AUC or AUROC)[47]. To train and validate the model we used tenfold cross validation. AUC results are presented as mean ± SD calculated over the tenfold validation sets. To test whether an AUC of logistic regression and ML models prediction could obtain similar results, ROC curve analysis was used to compare the 7-input variables selected for ML experiments of the different patient groups as potential morbidity and mortality clinical markers at 3 months. The statistical descriptive analysis was conducted in SPSS 25.0 (IBM, Chicago, IL) for Mac.

## Data availability
All data are available within the text of the manuscript. Further anonymized data could be made available to qualified investigators upon reasonable request.

### Acknowledgements
None.

### Author contributions
Drs. J.C., S.R.-Y., and R.I.-R. conceived the scientific idea, designed the experiments, assisted with statistical analysis, and provided input in the writing of the manuscript. Drs. C.F.-L., V.M.-A., and S.S.-G. performed machine learning analysis, and provided input in the writing of the manuscript. Drs. M.R.-Y., and I.L.-D., had major role in the acquisition of data. Neuroimaging study, interpreted the data, revised the manuscript. A.E.-G., T.S., F.C., P.H. provided discussions on the project throughout, interpreted data and also provided input in the writing of the manuscript. All authors reviewed the manuscript.

### Funding
This study was partially supported by grants from the Spanish Ministry of Science and Innovation (SAF2017-84267-R), Xunta de Galicia (Axencia Galega de Innovación (GAIN): IN607A2018/3), Instituto de Salud Carlos III (ISCIII) (PI17/00540, PI17/01103), Spanish Research Network on Cerebrovascular Diseases RETICS-INVICTUS PLUS (RD16/0019) and by the European Union FEDER program. T. Sobrino (CPII17/00027), F. Campos (CPII19/00020) are recipients of research contracts from the Miguel Servet Program (Instituto de Salud Carlos III). General Directorate of Culture, Education and University Management of Xunta de Galicia (ED431G/01,252 ED431D 2017/16), "Galician Network for Colorectal Cancer Research" (Ref. ED431D 2017/23), Competitive Reference Groups (ED431C 2018/49), Spanish Ministry of Economy and Competitiveness via funding of the unique installation BIOCAI (UNLC08-1E-002, UNLC13-13–3503), European Regional Development Funds (FEDER).

### Competing interests
The authors declare no competing interests.

### Additional information
**Supplementary Information** The online version contains supplementary material available at https://doi.org/10.1038/s41598-021-89434-7.

**Correspondence** and requests for materials should be addressed to S.R.-Y. or R.I.-R.

**Reprints and permissions information** is available at www.nature.com/reprints.

**Publisher's note** Springer Nature remains neutral with regard to jurisdictional claims in published maps and institutional affiliations.